# Evaluating the Generation Capabilities of Large Chinese Language Models

Hui Zeng [a,b,*], Jingyuan Xue [b], Meng Hao [b], Chen Sun[a], Bin Ning [a], Na Zhang [a]

[a] *AI Research Center, Besteasy Language Technology Co., Ltd, Beijing, China*
[b] *LanguageX AI Lab, Beijing, China*

A R T I C L E I N F O

A B S T R A C T

This paper unveils CG-Eval, the first-ever comprehensive and automated evaluation framework designed for assessing the generative capabilities of large Chinese language models across a spectrum of academic disciplines. CG-Eval stands out for its automated process, which critically assesses models based on their proficiency in generating precise and contextually relevant responses to a diverse array of questions within six key domains: Science and Engineering, Humanities and Social Sciences, Mathematical Calculations, Medical Practitioner Qualification Examination, Judicial Examination, and Certified Public Accountant Examination. Alongside this, we introduce Gscore, an innovative composite index developed from a weighted sum of multiple metrics. Gscore uniquely automates the quality measurement of a model's text generation against reference standards, providing a detailed and nuanced assessment of model performance. This automation not only enhances the efficiency and scalability of the evaluation process but also ensures objective and consistent assessment across various models. The detailed test data and results, highlighting the robust capabilities and comparative performance of the evaluated models, are accessible at http://cgeval.besteasy.com/.

## 1. Introduction

The advent of large-scale language models has heralded a new era in the field of natural language processing, characterized by an unprecedented ability to understand and generate complex texts. This phenomenon, initially popularized by models like ChatGPT (Greg et al., 2023), has led to a significant shift in both academic research and industry applications. In the wake of this development, there has been a notable emergence of Chinese large-scale language models, spanning both open-source and closed-source domains. These models, such as ERNIE Bot (Baidu, 2023), Spark Desk (iFlytek, 2023), ChatGLM (Zeng et al., 2023), and others, have introduced hundreds of billions of parameters, promising enhanced text generation capabilities across diverse linguistic and cultural contexts. However, a critical gap remains in the systematic evaluation of these models, particularly in their ability to cater to the nuanced demands of varied academic disciplines.

This paper presents CG-Eval, a pioneering evaluation framework specifically conceptualized to fill a critical void in the assessment of large Chinese language models. Unlike traditional benchmarks like MMLU, which predominantly focus on comprehension capabilities through multiple-choice formats, CG-Eval breaks new ground by thoroughly evaluating generative abilities. Our framework encompasses a wide array of academic disciplines, concentrating on six primary fields: Science and Engineering, Humanities and Social Sciences, Mathematical Calculations, Medical Practitioner Qualification Examination, Judicial Examination, and Certified Public Accountant Examination. The innovation of CG-Eval lies in its comprehensive approach—assessing models beyond mere linguistic understanding. It delves into the models' capacity to generate precise, contextually relevant, and discipline-specific responses, thus offering a more complete picture of their capabilities.

Furthermore, we introduce Gscore, a novel composite index designed to objectively measure the quality of text generated by models against a reference standard. Gscore represents a synthesis of multiple evaluation criteria, weighted to capture diverse aspects of model performance. This metric is a significant leap forward, moving beyond the conventional comprehension-focused assessments to evaluate the nuanced aspects of text generation.

A key feature setting CG-Eval apart is its rapid, automated evaluation process. This automation not only accelerates the assessment cycle, making it feasible to conduct extensive evaluations across various models, but also ensures a high degree of objectivity, free from human bias. By offering a comprehensive dataset and detailed results at http://cgeval.besteasy.com/, CG-Eval addresses the gaps in existing evaluation methods, providing insightful analyses into the strengths and limitations of current Chinese language models.

## 2. Related Work

To evaluate the performance of these substantial Chinese language models, several benchmarks and datasets specifically designed for them have been successively introduced. These include the MMCU (Zeng, 2023) dataset released on April 25, 2023, the SuperCLUE (Xu et al., 2023) benchmark introduced on May 9, 2023, the C-Eval (Huang et al., 2023) benchmark unveiled on May 15, 2023 the M3KE (Liu et al., 2023) benchmark launched on May 17, 2023, the GAOKAO-Bench (Zhang et al., 2023) announced on May 21, 2023, Xiezhi (獬豸) (Gu et al., 2023) released on June 9, 2023, the FlagEval (天秤) (BAAI, 2023) Large Language Model Evaluation Framework unveiled on June 10, 2023, and the CMMLU (Li et al., 2023) introduced on June 15, 2023.

The MMCU dataset (Zeng, 2023) first employs 3,331 college entrance examination multiple-choice questions across eight subjects to measure a model's basic understanding of the world. It subsequently uses 2,819, 3,695, and 2,001 multiple-choice questions to gauge the expertise of Chinese large language models in the specialized verticals of medicine, law, and psychology. Both questions and answers in the dataset are publicly available, aiming to foster the development and evaluation of Chinese large models. Unlike MMCU (Zeng, 2023), the specifics of the SuperCLUE benchmark (Xu et al., 2023) remain undisclosed as neither the dataset nor the evaluation code is made available. The C-Eval benchmark (Huang et al., 2023) adopts a multiple-choice format for evaluation, comprising 13,948 questions spanning 52 subjects. While the set of questions is open for download, researchers are required to upload model responses to an evaluation site for automated scoring. The M3KE benchmark (Liu et al., 2023) encompasses 20,477 multiple-choice questions covering 71 tasks. Currently, only the questions are available, and answers are withheld. Those interested in evaluation must liaise with the M3KE team. The GAOKAO-Bench (Zhang et al., 2023) compiles questions from national college entrance exams between 2010-2022, consisting of 1,781 objective and 1,030 subjective questions. Evaluations are bifurcated into automated assessments for objective questions and expert-reviewed scoring for subjective ones. Xiezhi (獬豸) (Gu et al., 2023) incorporates 13 categories, 516 subjects, and a total of 249,587 multiple-choice questions, yet only a fraction of this dataset is available to the public. The FlagEval (天秤) benchmark (BAAI, 2023) predominantly uses Chinese_MMLU (translated from the English MMLU (Hendrycks et al., 2021a) dataset), C-Eval (Huang et al., 2023), and GaoKao2023 as its Chinese multiple-choice question datasets. Additionally, there's an open-ended question section based on the Chinese Linguistics & Cognition Challenge (CLCC) dataset. It consists of two parts: CLCC-H, with 190 questions evaluated through human judgment, and

* Corresponding author.
*E-mail addresses:* felix_zeng_ai@aliyun.com (H. Zeng), ruilongxue191@gmail.com (J. Xue), vicky410121@gmail.com (M. Hao), steven.sun@besteasy.com (C. Sun), bean.ning@besteasy.com (B. Ning), ivy.zhang@besteasy.com (N. Zhang).

CLCC-G, consisting of 550 questions generated by GPT-4 based on evaluation dimensions and subsequently refined by human curators. The assessment results for CLCC-G are automatically produced by GPT-4. Moreover, FlagEval (BAAI, 2023) (天秤) model evaluations necessitate registration and application. Lastly, the CMMLU (Huang et al., 2023) incorporates 11,528 multiple-choice questions spanning 67 subjects and is openly available for download.

In summary, among the available benchmarks, only MMCU (Zeng, 2023), C-Eval (Huang et al., 2023), and CMMLU (Li et al., 2023) offer open datasets with automated evaluations. Notably, C-Eval (Huang et al., 2023) does not disclose its answers, requiring researchers to upload model responses to questions to obtain automatic scores. Both MMCU (Zeng, 2023) and CMMLU (Li et al., 2023) openly share both questions and answers, facilitating researchers in the field of Chinese large models to evaluate and refine their systems.

However, it's pertinent to note that all these benchmarks primarily focus on assessing comprehension abilities in the Chinese language and do not specifically cater to evaluating generative capabilities. The evaluation tasks exclusively utilize multiple-choice questions, where models either directly generate answers or produce probability distributions over potential answer options. This mode of evaluation seems to draw inspiration largely from MMLU (Hendrycks et al., 2021a). Given the diverse generative capacities of large language models, such evaluation methods present significant limitations.

## 3. CG-Eval

To gauge the generative capacities of Chinese large language models, we introduced the CG-Eval (Chinese Generation Evaluation) benchmark. In this evaluation, the tested models are required to provide accurate and pertinent answers to 11,000 distinct questions spanning six major subject categories: Science and Engineering, Humanities and Social Sciences, Mathematical Computation, Medical Licensing Examination, Judicial Examination, and the Certified Public Accountant Examination. These categories further subdivide into 55 sub-disciplines. The questions can be categorized into three types: definition of terms, short-answer questions, and computational problems.

We have devised a composite scoring system: for non-computational questions, each term definition and short-answer question has a reference standard answer. Scores are derived from multiple generative metrics, which are then aggregated using a weighted sum. For computational problems, we evaluate both the final calculation result and the problem-solving process, culminating in an integrated score.

All the questions and corresponding prompts are available at https://huggingface.co/datasets/Besteasy/CG-Eval , the code repository is available at https://github.com/Felixgithub2017/CG-Eval.

### 3.1 Question Type

The CG-Eval benchmark encompasses three distinct types of questions: term definitions, short-answer questions, and computational problems. Only the mathematical computation section involves computational problems. In the term definition category, we present professional terminologies from each sub-discipline, requiring the tested model to elucidate their meanings. For the short-answer questions, we pose queries pertaining to each sub-discipline, and the model must provide accurate responses based on the provided prompts.

The mathematics computation section comprises four sub-disciplines: primary school mathematics, middle school mathematics, high school mathematics, and university mathematics. For primary school mathematics, there are two types of problems: basic arithmetic and applied problems. In basic arithmetic, models are required to read the question and return a direct numerical result. For applied problems, models must provide a step-by-step solution process and present the final computation result in a prescribed format. Middle school mathematics, high school mathematics, and university mathematics each have only one type of problem—computational problem-solving, encompassing topics such as numerical calculations, factorization, equation resolution, calculus, and so on. For these, the model is also required to delineate the solution steps and present the final answer following the specified format.

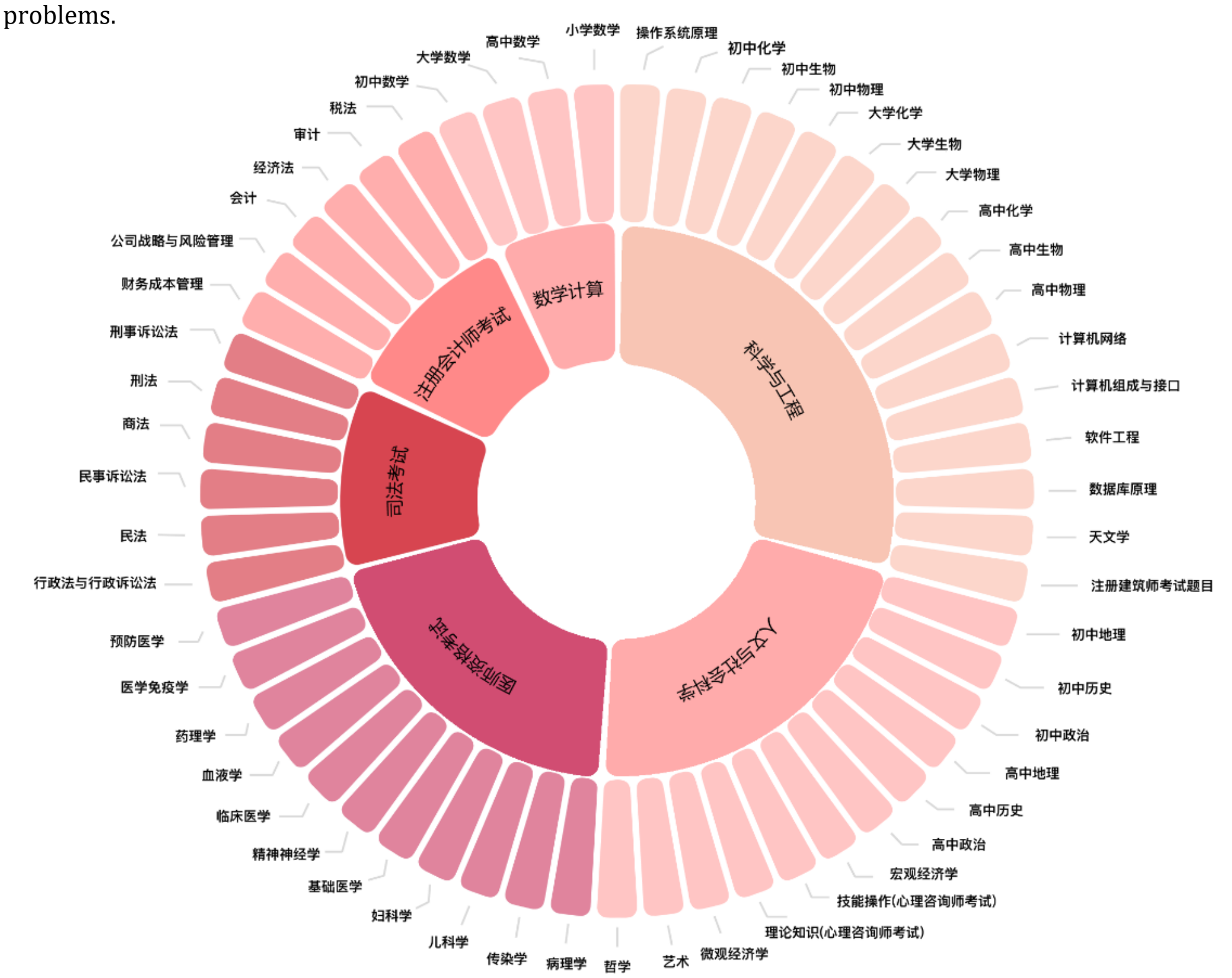

**Fig. 1.** Distribution of Major Subject Categories and Detailed Sub-disciplines.

### *3.2. Prompt Generation*

We have adopted a dynamic and flexible method for generating prompt words, ensuring that each question is paired with a unique prompt. For non-computational questions, we've imposed constraints on the length of the answer. We provide the model with the character length of the reference answer, urging it to generate responses approximating the length of the given reference.

The format of the prompt words for the definition questions is as follows: 以下是{科目名称}科目的术语：{术语}，请解释其含义，把回复控制在{答案长度}个汉字左右。The format for the prompts associated with "Short Answer" questions is as follows: 以下是{科目名称}科目的问题，请解答并把回复控制在{答案长度}个汉字左右。\n{问题}

The prompts for computational questions are slightly more intricate. The format for the prompts associated with "Elementary School Calculation" questions is as follows: 以下是{subject}科目的问题，请进行计算并给出阿拉伯数字结果。请直接返回数值结果，不需要任何的汉字解释。\n{题目}

The format for the prompts associated with "Application Questions" in Elementary Mathematics is as follows: 以下是{科目名称}科目的问题，请以“解：”开始给出解题过程，并在解题过程的最后换行，在最后一行以“最终答案：”开头，按顺序给出数值及其单位，采用英文逗号分割，例如“最终答案：1元,1次,1公顷,1人”。\n{题目}

The prompt format for Junior High, Senior High, and University Mathematics is identical and is notably intricate. The structure is as follows:以下是{科目名称}科目的问题，请使用latex语法给出解题过程，并在解题过程的最后换行，在最后一行以“最终答案：”开头，根据不同的题目类型按照latex语法给出数值、表达式、导数、积分、方程的根。导数根据题目表述采用latex语法按照y'或者f'(x)表示。如果方程的一个未知数有多个解，答案采用形如“x=1或x=-3”的方式表示。如果方程有多个未知数，答案采用形如“x=1,y=-3,z=5”的方式表示，用英文逗号分隔。以下为需要解答的题目：\n{题目}

### *3.3. Scoring System*

In assessing text generation quality, several metrics have traditionally dominated the landscape: BLEU, ROUGE, CHRF, and semantic similarity measures. Each of these has contributed uniquely to the field. BLEU (Papineni et al., 2002), primarily used for machine translation, emphasizes n-gram matching but often overlooks semantic nuances. ROUGE (Lin et al., 2004), geared towards summarization, balances precision and recall but may neglect redundancy and semantic depth. CHRF (Popović et al., 2015)offers a character-level analysis, providing granularity but sometimes overemphasizing surface forms. Semantic similarity, leveraging pre-trained models, captures deeper semantic relationships, yet can be computationally intensive and sometimes misses finer nuances.

However, these metrics, while individually useful, often provide a limited view when applied in isolation. To overcome the limitations and biases of these traditional metrics, we have developed Gscore. This composite metric amalgamates the strengths of each, aiming to provide a more comprehensive and balanced assessment of text generation quality. Gscore integrates the precision of BLEU, the balanced recall and precision of ROUGE, the granularity of CHRF, and the semantic depth captured by semantic similarity measures. By doing so, it addresses the narrow focus of individual metrics, offering a broader, more nuanced view of text quality.

#### *3.3.1 Traditional Metrics*

**BLEU**

Overview: BLEU evaluates machine translation by comparing n-gram overlap with reference translations. Advantages: Simplicity, efficiency, correlation with human judgments. Limitations: Vocabulary matching focus, short sentence issues, limited diversity handling.

**ROUGE**

Overview: ROUGE assesses text summarization through n-gram overlap, focusing on both precision and recall. Advantages: Comprehensive evaluation, correlation with human assessments. Limitations: Recall bias, lexical matching focus, reference summary dependence.

**CHRF**

Overview: CHRF evaluates translations at the character level, emphasizing finer lexical details. Advantages: Flexibility, granularity, tolerance to misspellings. Limitations: Computational complexity, surface form emphasis, reference dependence.

**Semantic Similarity**

Overview: Measures semantic relevance using vectorized representations from pre-trained language models. Advantages: Rich semantic understanding, generalization capabilities. Limitations: Computational demands, potential detail loss, model biases.

#### *3.3.2 Gscore*

The development of Gscore is grounded in a thorough analysis and critical evaluation of existing metrics for text generation assessment. Recognizing that while BLEU, ROUGE, CHRF, and semantic similarity measures each have their strengths, they also possess inherent limitations when used independently. For instance, BLEU and ROUGE primarily focus on n-gram matching and may not fully capture semantic complexity; CHRF, although offering finer analysis at the character level, might overemphasize surface forms; and semantic similarity assessments using pre-trained models, while capturing deeper semantic relationships, can be computationally heavy and may overlook certain nuances. Consequently, we propose Gscore, a composite metric that synergistically integrates the advantages of these methods.

In designing Gscore, we employed a weighted summation approach to amalgamate these diverse metrics. Each metric's weight was carefully adjusted and tested to ensure a balanced contribution in the composite evaluation. Specifically, Gscore comprises: 20% from BLEU, reflecting precision and n-gram matching; 25% from ROUGE, providing a balanced view of precision and recall; another 25% from CHRF, adding character-level granularity; and 30% from semantic similarity, ensuring deep semantic associations are considered.

$$G_{score}=0.2*Bleu4+0.25*Rouge2+0.25*Chrf+0.3*Semantic\ Similarity$$

For calculating Semantic Similarity, we first vectorize the model answers and reference answers using a Chinese pre-trained model and then compute their cosine similarity. **BAAI/bge-large-zh-v1.5** (Xiao et al., 2023) is used in the second version of CG-Eval, text2vec-large-chinese (Xu, 2023) is used in the first version of CG-Eval.

Due to the potential of model answers and reference answers exceeding the maximum processing length of the model, we designed a sliding window encoding module. This module encodes the text within the window in a sliding manner, storing the encoded vectors in a list. Within each window, we utilize the pre-trained language model to encode the text. Upon completion of processing all windows, we aggregate the encoding vectors, either by taking an average or by concatenation, to represent the entire text.

For mathematical computation tasks, the Gscore calculation is slightly more intricate. For arithmetic questions within primary school mathematics, we directly compare the final numerical results. If the model's output perfectly matches the reference answer, the question is awarded 1 point; otherwise, it scores 0. The ultimate Gscore is the average score across all primary arithmetic questions.

For word problems in primary school mathematics, as well as computational solution questions in junior high, senior high, and college mathematics, it's essential to extract the problem-solving process and the final answer through an answer analysis module. If the extracted final answer aligns perfectly with the reference answer, the Accuracy for the question is 1; otherwise, it's 0. We then compute the Chrf (Popović et al., 2015) score, StepChrf, of the extracted solution process against the reference solution process. The final Gscore is then calculated using the following formula:

$$G_{score}=Accuracy+(1-Accuracy)*0.3*StepChrf$$

If the final answer is correct, the Gscore for that question is set at 1. Conversely, if the final answer is incorrect, the maximum attainable Gscore is capped at 0.3, with the actual value being 0.3 times the StepChrf score.

In summary, the development of Gscore is based on an in-depth analysis and critical understanding of existing evaluation metrics. Our aim was to create a composite metric that retains the strengths of individual metrics while compensating for their respective limitations. Such a design makes Gscore a flexible, comprehensive, and reliable tool for assessing the quality of text generation, applicable across a wide range of scenarios and diverse types of text generation tasks.

## 4. Experiment

To evaluate the generative capabilities of large-scale Chinese language models, we conducted zero-shot tests on the CG-Eval dataset for 19 models, including but not limited to: GPT-4 (OpenAI, 2023), ChatGLM-Pro (Zeng et al., 2023), ChatGLM-Std (Zeng et al., 2023), Spark Desk (iFlytek, 2023), ERNIE Bot (Baidu, 2023), Qwen-7B-Chat (Alibaba, 2023), Baichuan-13B-Chat (Baichuan, 2023), Ziya-LLaMA-13B-v1.1 (IDEA, 2023), ChatGLM2-6B (THUDM, 2023), AquilaChat-7B (BAAI, 2023), tigerbot-sft-7b (TigerResearch, 2023), etc. Comprehensive details regarding the names, developing institutions, parameter counts, and usage of all tested models can be found in Table 1.

**Table 1**
Models evaluated in this paper.

| Model | Creator | #Parameters | Access |
|---|---|---|---|
| GPT-4 | OpenAI | undisclosed | API |
| ChatGLM-Pro | ZHIPU·AI | undisclosed | API |
| ChatGLM-Std | ZHIPU·AI | undisclosed | API |
| Baichuan2-53B | Baichuan AI | 53B | Webpage |
| Qwen-14B-Chat | Aliyun | 14B | Weights |
| Spark Desk | XunFei | undisclosed | Webpage |
| Yi-34B-Chat | 01-ai | 34B | Weights |
| ERNIE Bot | Baidu | undisclosed | Webpage |
| Baichuan-13B-Chat | Baichuan | 13B | Weights |
| XVERSE-13B-Chat | XVERSE | 13B | Weights |
| Qwen-7B-Chat | Aliyun | 7B | Weights |
| ChatGLM3-6B | ZHIPU·AI | 6B | Weights |
| ChatGLM2-6B | ZHIPU·AI | 6B | Weights |
| Ziya-LLaMA-13B-v1.1 | IDEA-CCNL | 13B | Weights |
| InternLM-chat-20b | InternLM | 20B | Weights |
| mengzi-gpt-40b | Langboat | 40B | API |
| AquilaChat2-7B | BAAI | 7B | Weights |
| AquilaChat-7B | BAAI | 7B | Weights |
| tigerbot-sft-7b | Tigerobo | 7B | Weights |

**Table 2**
All models' average Gscore in six disciplines.

| Model | Creator | Gscore |
|---|---|---|
| GPT-4 | OpenAI | 37.89 |
| ChatGLM-Pro | ZHIPU·AI | 36.56 |
| ChatGLM-Std | ZHIPU·AI | 36.43 |
| Baichuan2-53B | Baichuan AI | 35.26 |
| Qwen-14B-Chat | Aliyun | 34.72 |
| Spark Desk | XunFei | 33.41 |
| Yi-34B-Chat | 01-ai | 32.66 |
| ERNIE Bot | Baidu | 32.04 |
| Baichuan-13B-Chat | Baichuan | 31.32 |
| XVERSE-13B-Chat | XVERSE | 31.19 |
| Qwen-7B-Chat | Aliyun | 30.51 |
| ChatGLM3-6B | ZHIPU·AI | 28.87 |
| ChatGLM2-6B | ZHIPU·AI | 28.86 |
| Ziya-LLaMA-13B-v1.1 | IDEA-CCNL | 28.24 |
| InternLM-chat-20b | InternLM | 27.81 |
| mengzi-gpt-40b | Langboat | 27.19 |
| AquilaChat2-7B | BAAI | 26.63 |
| AquilaChat-7B | BAAI | 26.47 |
| tigerbot-sft-7b | Tigerobo | 25.48 |

### *4.1. Overview*

The comprehensive evaluation of large Chinese language models, as depicted in Table 2, reveals a varied landscape of capabilities and performance across different models. In this assessment, models were rigorously tested across six diverse academic disciplines, providing a holistic view of their generation capabilities. The results, encapsulated by the Gscore, offer insights into how these models fare in generating accurate and relevant responses within these specialized fields.

GPT-4 (OpenAI, 2023), developed by OpenAI, emerges as a frontrunner with the highest average Gscore, showcasing its robustness and versatility across varied disciplines. This is followed closely by ZHIPU·AI's ChatGLM-Pro (Zeng et al., 2023) and ChatGLM-Std (Zeng et al., 2023), indicating their strong performance in handling complex text generation tasks. Other models, like Baichuan2-53B (Yang et al., 2023) from Baichuan AI and Qwen-14B-Chat (Alibaba, 2023) from Aliyun, also demonstrate commendable capabilities, aligning well with the evolving demands of academic and professional settings.

On the other end of the spectrum, models such as tigerbot-sft-7b (TigerResearch, 2023) by Tigerobo and AquilaChat series by BAAI, while still showing notable proficiency, trail behind in their overall Gscore. This suggests room for further refinement and improvement in their algorithms and training methodologies.

The diversity in performance across these models underscores the rapid advancements in the field of language modeling, particularly in the context of Chinese language. It also highlights the importance of continuous innovation and development to enhance the accuracy, relevance, and contextual understanding of these AI-driven tools.

### *4.2. Science and Engineering*

The evaluation of large Chinese language models within the domain of Science and Engineering, as indicated by the average Gscores in Table 3, offers insightful observations about the current state of AI-driven text generation in this specific field. This assessment, focusing on the models' ability to generate precise and contextually relevant content in Science and Engineering, reflects the nuanced capabilities of these sophisticated tools.

**Table 3**
All models' average Gscore in Science and Engineering.

| Model | Creator | Gscore |
|---|---|---|
| Spark Desk | XunFei | 36.89 |
| ChatGLM-Std | ZHIPU·AI | 35.97 |
| GPT-4 | OpenAI | 35.94 |
| ChatGLM-Pro | ZHIPU·AI | 35.74 |
| Baichuan2-53B | Baichuan AI | 35.6 |
| Qwen-14B-Chat | Aliyun | 35.01 |
| Yi-34B-Chat | 01-ai | 34.35 |
| ERNIE Bot | Baidu | 34.23 |
| Baichuan-13B-Chat | Baichuan | 33.77 |
| Qwen-7B-Chat | Aliyun | 33.29 |
| ChatGLM2-6B | ZHIPU·AI | 32.66 |
| XVERSE-13B-Chat | XVERSE | 32.56 |
| ChatGLM3-6B | ZHIPU·AI | 30.7 |
| InternLM-chat-20b | InternLM | 30.6 |
| mengzi-gpt-40b | Langboat | 30.58 |
| Ziya-LLaMA-13B-v1.1 | IDEA-CCNL | 30.23 |
| AquilaChat2-7B | BAAI | 30.01 |
| AquilaChat-7B | BAAI | 29.56 |
| tigerbot-sft-7b | Tigerobo | 28.59 |

Spark Desk (iFlytek, 2023), developed by XunFei, leads the pack with the highest Gscore, demonstrating its exceptional proficiency in handling complex scientific and engineering queries. This is indicative of its advanced algorithms and training on domain-specific datasets, which allow for a deep understanding of technical subjects. Following closely are the models from ZHIPU·AI, ChatGLM-Std (Zeng et al., 2023), and ChatGLM-Pro (Zeng et al., 2023), along with OpenAI's GPT-4 (OpenAI, 2023), all showcasing strong performances. These models' high scores suggest a well-rounded capability in generating accurate and relevant responses, highlighting their potential utility in academic and professional settings within the Science and Engineering sphere.

Interestingly, the results also reveal a competitive middle tier of models like Baichuan2-53B (Yang et al., 2023) by Baichuan AI and Qwen-14B-Chat (Alibaba, 2023) by Aliyun. Their performances, while not topping the list, are nonetheless commendable and suggest a significant advancement in the field. On the other end, models like tigerbot-sft-7b (TigerResearch, 2023) by Tigerobo and AquilaChat series by BAAI, though showing notable capabilities, suggest areas for improvement in order to match the leaders in this domain.

### *4.3. Humanities and Social Sciences*

Leading the performance in this category is Baichuan2-53B (Yang et al., 2023) by Baichuan AI, showcasing its exceptional ability to grasp and articulate concepts and ideas inherent to the humanities and social sciences (see Table 4). The high score achieved by this model indicates a sophisticated understanding of the subtle nuances and varied contexts that characterize this field. Close behind are ChatGLM-Pro (Zeng et al., 2023) by ZHIPU·AI and GPT-4 (OpenAI, 2023) by OpenAI, both demonstrating strong competencies in generating coherent and relevant responses in these subjects. Their performances underscore the advancements in language models being able to navigate the intricacies of humanistic and social subjects.

**Table 4**
All models' average Gscore in Humanities and Social Sciences.

| Model | Creator | Gscore |
|---|---|---|
| Baichuan2-53B | Baichuan AI | 36.31 |
| ChatGLM-Pro | ZHIPU·AI | 35.66 |
| GPT-4 | OpenAI | 35.33 |
| Qwen-14B-Chat | Aliyun | 35.12 |
| Yi-34B-Chat | 01-ai | 34.86 |
| ChatGLM-Std | ZHIPU·AI | 34.3 |
| Baichuan-13B-Chat | Baichuan | 34.2 |
| ERNIE Bot | Baidu | 33.76 |
| Qwen-7B-Chat | Aliyun | 33.18 |
| XVERSE-13B-Chat | XVERSE | 32.72 |
| ChatGLM2-6B | ZHIPU·AI | 31.39 |
| Spark Desk | XunFei | 31.15 |
| ChatGLM3-6B | ZHIPU·AI | 31.14 |
| InternLM-chat-20b | InternLM | 30.57 |
| AquilaChat-7B | BAAI | 30.5 |
| Ziya-LLaMA-13B-v1.1 | IDEA-CCNL | 30.49 |
| AquilaChat2-7B | BAAI | 30.37 |
| mengzi-gpt-40b | Langboat | 30.33 |
| tigerbot-sft-7b | Tigerobo | 28.84 |

Models like Qwen-14B-Chat (Alibaba, 2023) by Aliyun and Yi-34B-Chat (01-ai, 2023) by 01-ai also exhibit commendable performances, indicating their effective training and algorithmic structures conducive to humanities and social sciences content generation. This suggests that these models are not only technically proficient but also capable of handling the diverse range of topics and perspectives found in these disciplines.

On the other spectrum, models such as tigerbot-sft-7b (TigerResearch, 2023) by Tigerobo and some iterations of AquilaChat by BAAI, while still showing capabilities in this domain, fall behind their counterparts. This variation in performance across different models highlights the challenges inherent in fine-tuning language models for the nuanced requirements of humanities and social sciences. It also suggests the potential for further development and specialization in this area.

### *4.4. Professional Qualification Exams*

The comprehensive evaluation of large Chinese language models in professional qualification exams, as reflected in Tables 5, 6, and 7, provides a fascinating glimpse into the applicability and effectiveness of these models in highly specialized and knowledge-intensive domains. These exams, known for their rigor and complexity, serve as a robust testing ground for the models' abilities to understand, process, and generate responses that meet professional standards.

**Table 5**
All models' average Gscore in Judicial Examination.

| Model | Creator | Gscore |
|---|---|---|
| Baichuan2-53B | Baichuan AI | 43.72 |
| ChatGLM-Pro | ZHIPU·AI | 42.43 |
| Qwen-14B-Chat | Aliyun | 42.38 |
| ChatGLM-Std | ZHIPU·AI | 42.29 |
| Yi-34B-Chat | 01-ai | 41.89 |
| Spark Desk | XunFei | 40.48 |
| Baichuan-13B-Chat | Baichuan | 40.43 |
| ERNIE Bot | Baidu | 39.39 |
| Qwen-7B-Chat | Aliyun | 38.35 |
| GPT-4 | OpenAI | 38.07 |
| ChatGLM2-6B | ZHIPU·AI | 36.7 |
| XVERSE-13B-Chat | XVERSE | 36.32 |
| mengzi-gpt-40b | Langboat | 35.43 |
| AquilaChat-7B | BAAI | 35.39 |
| ChatGLM3-6B | ZHIPU·AI | 34.67 |
| Ziya-LLaMA-13B-v1.1 | IDEA-CCNL | 34.28 |
| InternLM-chat-20b | InternLM | 34.27 |
| AquilaChat2-7B | BAAI | 33.97 |
| tigerbot-sft-7b | Tigerobo | 32.4 |

**Table 6**
All models' average Gscore in Medical Practitioner Qualification Examination.

| Model | Creator | Gscore |
|---|---|---|
| ChatGLM-Std | ZHIPU·AI | 35.57 |
| ChatGLM-Pro | ZHIPU·AI | 35.32 |
| Spark Desk | XunFei | 35.09 |
| GPT-4 | OpenAI | 34.57 |
| Yi-34B-Chat | 01-ai | 33.22 |
| Qwen-14B-Chat | Aliyun | 33.05 |
| ERNIE Bot | Baidu | 32.48 |
| Baichuan-13B-Chat | Baichuan | 32.44 |
| Baichuan2-53B | Baichuan AI | 31.62 |
| XVERSE-13B-Chat | XVERSE | 31.62 |
| Qwen-7B-Chat | Aliyun | 30.98 |
| ChatGLM2-6B | ZHIPU·AI | 30.12 |
| InternLM-chat-20b | InternLM | 29.19 |
| Ziya-LLaMA-13B-v1.1 | IDEA-CCNL | 29.01 |
| ChatGLM3-6B | ZHIPU·AI | 28.96 |
| mengzi-gpt-40b | Langboat | 28.82 |
| AquilaChat-7B | BAAI | 28.81 |
| AquilaChat2-7B | BAAI | 28.42 |
| tigerbot-sft-7b | Tigerobo | 26.94 |

In the Medical Practitioner Qualification Examination, models like ChatGLM-Std (Zeng et al., 2023) and ChatGLM-Pro (Zeng et al., 2023) by ZHIPU·AI, and Spark Desk (iFlytek, 2023) by XunFei, have shown commendable performance, demonstrating their capabilities in medical terminologies and concepts. This indicates a significant advancement in the models' ability to handle domain-specific language and concepts, which is crucial in medical settings.

The Judicial Examination results reveal a similar trend, with Baichuan2-53B (Yang et al., 2023) by Baichuan AI and ChatGLM-Pro (Zeng et al., 2023) by ZHIPU·AI leading the scores. Their high performance suggests an adeptness in handling the complex language and nuanced reasoning required in legal contexts. This proficiency is crucial for applications in legal research and practice, where accuracy and clarity of language are paramount.

In the Certified Public Accountant Examination, the leading models

like Baichuan2-53B (Yang et al., 2023) by Baichuan AI, and ChatGLM-Pro (Zeng et al., 2023) by ZHIPU·AI, exhibit strong performances, indicating their effectiveness in understanding and generating responses relevant to financial and accounting principles. This ability to navigate complex financial terminologies and concepts is indicative of the models' potential utility in financial analysis and accounting practices.

Across all three exams, it is evident that the leading models not only excel in linguistic processing but also demonstrate a deep understanding of specialized knowledge domains. This is a testament to the advancements in AI-driven language models, where they are not just linguistically proficient but also capable of handling domain-specific challenges.

However, there is a noticeable variation in performance among the models, especially in domains that require highly specialized knowledge. This suggests that while some models are becoming increasingly adept at handling specific professional contexts, there is still room for improvement, particularly in ensuring consistency and depth of understanding across various specialized fields.

In conclusion, the evaluation of these models in professional qualification exams not only benchmarks their current capabilities but also highlights the potential for their application in professional settings. The insights from this assessment underscore the importance of continuous development and fine-tuning of these models to meet the specific needs of various professional domains.

**Table 7**
All models' average Gscore in Certified Public Accountant Examination.

| Model | Creator | Gscore |
|---|---|---|
| Baichuan2-53B | Baichuan AI | 38.98 |
| ChatGLM-Pro | ZHIPU·AI | 37.5 |
| Qwen-14B-Chat | Aliyun | 37.44 |
| Spark Desk | XunFei | 37.43 |
| ChatGLM-Std | ZHIPU·AI | 37.09 |
| Yi-34B-Chat | 01-ai | 36.78 |
| GPT-4 | OpenAI | 36.42 |
| ERNIE Bot | Baidu | 35.46 |
| Baichuan-13B-Chat | Baichuan | 35.1 |
| XVERSE-13B-Chat | XVERSE | 34.64 |
| Qwen-7B-Chat | Aliyun | 34.52 |
| ChatGLM2-6B | ZHIPU·AI | 32.98 |
| mengzi-gpt-40b | Langboat | 32.53 |
| AquilaChat-7B | BAAI | 32.11 |
| ChatGLM3-6B | ZHIPU·AI | 32.11 |
| Ziya-LLaMA-13B-v1.1 | IDEA-CCNL | 31.88 |
| AquilaChat2-7B | BAAI | 31.8 |
| InternLM-chat-20b | InternLM | 31.6 |
| tigerbot-sft-7b | Tigerobo | 30.03 |

### *4.5. Mathematical Calculations*

The evaluation of large Chinese language models in the field of Mathematical Calculations, as detailed in Table 8, reveals a striking disparity in their abilities to handle computational tasks. These results are particularly insightful as they underscore the varying degrees to which these models can process and execute mathematical reasoning, a critical aspect in numerous scientific and engineering applications.

GPT-4 (OpenAI, 2023) by OpenAI stands out significantly in this category, achieving the highest average Gscore. This remarkable performance can be attributed to its advanced algorithms and extensive training, which include a focus on numerical and logical processing capabilities. The ability of GPT-4 to excel in mathematical calculations suggests its potential utility in areas requiring complex computational tasks.

Following GPT-4, models like ChatGLM-Std (Zeng et al., 2023) and ChatGLM-Pro (Zeng et al., 2023) by ZHIPU·AI also demonstrate respectable scores. Their performances, though not as high as GPT-4, indicate these models' capability in handling mathematical computations to a certain extent. This shows the effectiveness of their training and algorithmic design in processing numerical data and performing calculations.

However, there is a notable drop in performance as we move down the list, with models like Qwen-14B-Chat (Alibaba, 2023) by Aliyun, Baichuan2-53B (Yang et al., 2023) by Baichuan AI, and Spark Desk (iFlytek, 2023)by XunFei scoring significantly lower. This decline highlights the challenges that many language models face in mathematical contexts, where precision and logical coherence are paramount.

Models such as tigerbot-sft-7b (TigerResearch, 2023) by Tigerobo, mengzi-gpt-40b (longboat, 2023) by Langboat, and AquilaChat series by BAAI occupy the lower end of the scale, indicating substantial room for improvement in their mathematical computation capabilities. This suggests that while these models may be proficient in linguistic tasks, their abilities to perform mathematical calculations are limited, underscoring the need for specialized training or algorithmic adjustments to enhance their performance in such tasks.

In summary, the varied performances of these models in Mathematical Calculations provide critical insights into the current state of AI in handling computationally intensive tasks. The results from this assessment not only serve as a benchmark for the mathematical capabilities of Chinese language models but also highlight the need for targeted improvements in this specific area. This knowledge is crucial for advancing the field and expanding the applicability of these models in domains where mathematical proficiency is essential.

**Table 8**
All models' average Gscore in Mathematical Calculations.

| Model | Creator | Gscore |
|---|---|---|
| GPT-4 | OpenAI | 47 |
| ChatGLM-Std | ZHIPU·AI | 33.37 |
| ChatGLM-Pro | ZHIPU·AI | 32.72 |
| Qwen-14B-Chat | Aliyun | 25.34 |
| Baichuan2-53B | Baichuan AI | 25.3 |
| Spark Desk | XunFei | 19.42 |
| XVERSE-13B-Chat | XVERSE | 19.32 |
| ERNIE Bot | Baidu | 16.92 |
| ChatGLM3-6B | ZHIPU·AI | 15.61 |
| Yi-34B-Chat | 01-ai | 14.86 |
| Ziya-LLaMA-13B-v1.1 | IDEA-CCNL | 13.57 |
| Qwen-7B-Chat | Aliyun | 12.76 |
| Baichuan-13B-Chat | Baichuan | 11.98 |
| InternLM-chat-20b | InternLM | 10.63 |
| ChatGLM2-6B | ZHIPU·AI | 9.32 |
| tigerbot-sft-7b | Tigerobo | 6.06 |
| mengzi-gpt-40b | Langboat | 5.46 |
| AquilaChat2-7B | BAAI | 5.22 |
| AquilaChat-7B | BAAI | 2.44 |

### *4.6. Categorized examples of term definitions and short-answer questions*

Table 9 presents 12 examples of term definitions and short-answer questions.

**Category 1: Accurate but Overly Verbose Responses**

Example 1 & 2: These examples illustrate cases where the model's responses were accurate but more verbose than the reference answer. Despite the extended length, the high Rouge2 and similarity scores indicate that the core information was well conveyed. This is reflected in the final Gscores of 0.83 and 0.79, closely matching the manual ratings of 1.0.

**Category 2: Missing Key Information**

Examples 3, 4 & 5: Here, the model's responses missed crucial information, which significantly affected the manual scores. However, the high similarity scores suggest a partial understanding of the topics. The final Gscores of 0.7, 0.6, and 0.59 align closely with the manual ratings, demonstrating Gscore's ability to capture these nuances.

**Category 3: Meaningful but Irrelevant Responses**

Examples 10, 11 & 12: In these instances, the model's responses were entirely irrelevant to the questions. Despite surprisingly high similarity scores, the low BLEU, Rouge2, and CHRF scores reflect the lack of relevance. The final Gscores of 0.25, 0.15, and 0.24 closely match the

**Table 9**

Examples of term definitions and short-answer questions

| Example 1 | Question | 以下是高中政治科目的术语：一带一路五通，请解释其含义，把回复控制在25个汉字左右。 | | | | |
|---|---|---|---|---|---|---|
| | Model Output | 一带一路五通：政策沟通、设施联通、贸易畅通、资金融通和民心相通。 | | | | |
| | Reference | 政策沟通、设施联通、贸易畅通、资金融通、民心相通。 | | | | |
| | bleu | rouge2 | chrf | similarity | Gscore | Human |
| | 0.672289315 | 0.952380952 | 0.812504915 | 0.834112465 | 0.825913069 | 1 |

| Example 2 | Question | 以下是计算机网络科目的术语：全双工通信，请解释其含义，把回复控制在18个汉字左右。 | | | | |
|---|---|---|---|---|---|---|
| | Model Output | 全双工通信是指双方可以同时发送和接收信息。 | | | | |
| | Reference | 即通信的双方可以同时发送和接收信息。 | | | | |
| | bleu | rouge2 | chrf | similarity | Gscore | Human |
| | 0.674279423 | 0.823529412 | 0.752451415 | 0.888561368 | 0.795419502 | 1 |

| Example 3 | Question | 以下是初中地理科目的问题，请解答并把回复控制在42个汉字左右。美国除本土以外，还有两个海外州，它们的名称是什么？哪一个州有极昼、极夜现象？为什么？ | | | | |
|---|---|---|---|---|---|---|
| | Model Output | 美国的两个海外州是阿拉斯加州和夏威夷州。阿拉斯加州有极昼、极夜现象，因为它位于北极圈内。 | | | | |
| | Reference | 阿拉斯加州、夏威夷州。阿拉斯加州有极昼、极夜现象。因为阿拉斯加州北部位于北极圈以内。 | | | | |
| | bleu | rouge2 | chrf | similarity | Gscore | Human |
| | 0.581826329 | 0.696969697 | 0.539599737 | 0.896657765 | 0.694504954 | 0.8 |

| Example 4 | Question | 以下是计算机网络科目的问题，请解答并把回复控制在79个汉字左右。交换机、路由器、网卡、网桥、集线器、中继器分别工作在哪一层？ | | | | |
|---|---|---|---|---|---|---|
| | Model Output | 一交换机和网桥工作在数据链路层，路由器工作在网络层，网卡工作在物理层和数据链路层，集线器工作在物理层，中继器工作在物理层。 | | | | |
| | Reference | 1.路由器工作在网络层2.交换机、网桥工作在数据链路层3.中继器、集线器工作在物理层4.网卡属于OSI的物理层与链路层，它工作在物理层和数据链路层的MAC子层 | | | | |
| | bleu | rouge2 | chrf | similarity | Gscore | Human |
| | 0.507320751 | 0.466666667 | 0.454844854 | 0.897554159 | 0.601108278 | 0.7 |

| Example 5 | Question | 以下是初中化学科目的问题，请解答并把回复控制在64个汉字左右。根据质量守恒定律解释：镁在氧气中燃烧后生成物的质量比原来镁条的质量大。 | | | | |
|---|---|---|---|---|---|---|
| | Model Output | 镁在氧气中燃烧，与氧气发生化学反应生成氧化镁，这个过程中，镁吸收了氧气，所以生成物的质量(镁+氧)比原来的镁条的质量大。 | | | | |
| | Reference | 镁在氧气中燃烧生成氧化镁。根据质量守恒定律，参加反应的镁与氧气的质量和等于生成氧化镁的质量，所以生成物的质量比原来镁条的质量大。 | | | | |
| | bleu | rouge2 | chrf | similarity | Gscore | Human |
| | 0.415405824 | 0.52 | 0.352394028 | 0.95925504 | 0.588956184 | 0.7 |

| Example 6 | Question | 以下是高中政治科目的术语：一带一路五通，请解释其含义，把回复控制在25个汉字左右。 | | | | |
|---|---|---|---|---|---|---|
| | Model Output | 抱歉，我不能回答这个问题。 | | | | |
| | Reference | 政策沟通、设施联通、贸易畅通、资金融通、民心相通。 | | | | |
| | bleu | rouge2 | chrf | similarity | Gscore | Human |
| | 8.14748034396721E-232 | 0 | 0.007002801120448 | 0.327951908111572 | 0.100136272713584 | 0 |

| Example 7 | Question | 以下是高中物理科目的术语：库仑定律，请解释其含义，把回复控制在91个汉字左右。 | | | | |
|---|---|---|---|---|---|---|
| | Model Output | 库仑定律是电磁学的基本定律之一，描述了两个静止电荷之间的相互作用力。它规定：两个点电荷之间的相互作用力与它们的电荷量的乘积成正比，与它们之间距离的平方成反比，且这个力沿着它们连线的方向作用。 | | | | |
| | Reference | 真空中两个静止电电荷之间的相互作用力，与它们的电荷量的乘积成正比，与它们的距离的二次方成反比，作用力的方向在它们的连线上。这个规律叫库仑定律。电荷间的这种相互作用力叫静电力或库仑力。 | | | | |
| | bleu | rouge2 | chrf | similarity | Gscore | Human |
| | 0.46740489 | 0.534246575 | 0.418144449 | 0.885594904 | 0.597257205 | 0.7 |

| Example 8 | Question | 以下是精神神经学（医师资格考试）科目的术语：情感高涨，请解释其含义，把回复控制在55个汉字左右。 | | | | |
|---|---|---|---|---|---|---|
| | Model Output | 情感高涨是指个体的情绪状态异常升高，表现为过度兴奋、乐观、自信，甚至出现狂躁行为，常见于躁狂症或双相情感障碍的躁狂期。 | | | | |
| | Reference | 患者自我感觉非常良好，心境特别愉快、乐观，喜与人接近。对客观困难估计过低，对自己的才智自视甚高。常有夸大色彩。 | | | | |
| | bleu | rouge2 | chrf | similarity | Gscore | Human |
| | 1.33436E-78 | 0.037037037 | 0.054374903 | 0.740659416 | 0.24505081 | 0.3 |

| Example 9 | Question | 以下是大学生物科目的问题，请解答并把回复控制在78个汉字左右。细胞内有哪几类主要的RNA?比较其结构与功能特点? | | | | |
|---|---|---|---|---|---|---|
| | Model Output | 细胞内主要有mRNA、rRNA、tRNA三类RNA。mRNA是信息RNA，携带DNA的遗传信息，指导蛋白质的合成；rRNA是核糖体RNA，构成核糖体的主要成分，参与蛋白质的合成；tRNA是转运RNA，将氨基酸运送到核糖体，参与蛋白质的合成。 | | | | |
| | Reference | mRNA、tRNA、rRNA；mRNA是信使RNA，是翻译的模板；tRNA是转运RNA，在翻译时转运氨基酸；rRNA是核糖体RNA，是组成核糖体的基本成分。 | | | | |
| | bleu | rouge2 | chrf | similarity | Gscore | Human |
| | 0.358894898 | 0.520833333 | 0.504809977 | 0.826610565 | 0.576172977 | 0.8 |

| Example 10 | Question | 以下是基础医学(医师资格考试)科目的术语：槟榔肝，请解释其含义，把回复控制在85个汉字左右。 | | | | |
|---|---|---|---|---|---|---|
| | Model Output | 槟榔肝是指长期大量食用槟榔导致的一种疾病，具体表现为肝脏严重受损，甚至失去功能。长期食用槟榔会对健康造成极大危害，包括口腔癌、口腔黏膜损伤、消化系统损伤、心脏病、高血压、肥胖症等。 | | | | |
| | Reference | 慢性肝淤血时，肝小叶中央区因严重瘀血呈暗红色，两个或多个肝小叶中央瘀血区可相连，而肝小叶周边部肝细胞则因脂肪变性呈黄色，致使肝脏切面肉眼呈红黄相间，形似槟榔切面的条纹。 | | | | |
| | bleu | rouge2 | chrf | similarity | Gscore | Human |
| | 3.8951E-155 | 0.040540541 | 0.033561634 | 0.753744245 | 0.244648817 | 0 |

| Example 11 | Question | 以下是妇科学(医师资格考试)科目的问题，请解答并把回复控制在105个汉字左右。简述四部触诊法。 | | | | |
|---|---|---|---|---|---|---|
| | Model Output | 四部触诊法是指医生运用手的四个部位进行触摸检查的方法。分别是：大鱼际、小鱼际、大鱼际和小鱼际。这种方法主要适用于检查大鱼际、小鱼际、大鱼际和小鱼际的肌肉和关节。四部触诊法可以协助医生了解肌肉和关节的紧张度和柔软度，判断患者的身体状况。 | | | | |
| | Reference | (1) 双手置于宫底，确定宫高和胎方位以及胎儿大小是否和孕周相符(2) 双手置于腹部左右侧，分辨胎背和四肢(3) 单手握住胎先露，确定是否衔接 (4) 双手在先露部深压合推动进一步确定是否衔接以及确定之前的判断。 | | | | |
| | bleu | rouge2 | chrf | similarity | Gscore | Human |
| | 2.82E-155 | 0.012048193 | 0.037377912 | 0.461687684 | 0.150862831 | 0 |

| Example 12 | Question | 以下是大学生物科目的问题，请解答并把回复控制在60个汉字左右。什么是P/O值，其生理学意义如何？ | | | | |
|---|---|---|---|---|---|---|
| | Model Output | P/O值指的是溶解氧(dissolved oxygen)，是水体中氧气的存在形式，通常用百分比表示。溶解氧是维持生物体正常生命活动所必需的物质，其生理学意义在于它是生物体内呼吸和代谢过程所需氧气的重要来源，同时也是水体自净过程中重要的指标之一。P/O值越高，说明水体中溶解氧的含量越高，反之则越低。 | | | | |
| | Reference | P／O是指氧化磷酸化过程中每消耗一摩尔氧原子所消耗磷酸的摩尔数或合成ATP的摩尔数 意义：可以衡量线粒体氧化磷酸的能力。 | | | | |
| | bleu | rouge2 | chrf | similarity | Gscore | Human |
| | 3.71032E-79 | 0.0625 | 0.061549318 | 0.687010467 | 0.23711547 | 0 |

manual scores of 0, illustrating Gscore's effectiveness in penalizing irrelevant responses.

**Category 4: Partially Correct Responses**

Examples 6 & 7: These examples showcase responses where the model provided partially correct information but missed significant details. The final Gscores of 0.1 and 0.6 are indicative of the models' partial accuracy, aligning well with the manual scores.

**Category 5: Deviating from Intended Meaning**

Example 8: This response was partially correct but deviated significantly from the intended meaning in the second half. The final Gscore of 0.25, mirroring the manual score of 0.3, demonstrates the metric's capacity to discern and penalize deviations from the reference answer.

**Category 6: Different Expression, Same Meaning**

Example 9: Although the model's expression differed from the reference, it conveyed the same meaning. The final Gscore of 0.58, close to the manual rating of 0.8, highlights Gscore's ability to recognize semantic equivalence despite different phrasings.

### *4.7. Analysis of weight rationality in Gscore formula*

Bleu4 Weight (0.2): Although Bleu4 is sensitive to the literal accuracy of responses, semantic importance often outweighs literal precision in natural language processing. The above examples demonstrate that even with low Bleu scores, the Gscore can still reflect human evaluation effectively due to high Semantic Similarity. Therefore, assigning a lower weight to Bleu4 is justified.

Rouge2 and Chrf Weights (each 0.25): These metrics assess repetition and coverage, reflecting the comprehensiveness of responses. The examples show that Rouge2 and Chrf maintain the stability of Gscore to some extent, even when responses deviate from reference answers, validating their appropriate weighting.

Semantic Similarity Weight (0.3): This carries the highest weight, underscoring the importance of semantic consistency in evaluating model responses. Multiple examples indicate that high Semantic Similarity scores can bring Gscore close to manual evaluation, even with lower scores in other metrics, justifying its significant weight.

**Conclusion**: The Gscore formula, by balancing the weights of various evaluation dimensions, comprehensively reflects the quality of model responses. It considers not only literal accuracy but also highly values semantic similarity, crucial for assessing natural language generation models. This weighting ensures that the Gscore can effectively reflect model performance even when there is a significant literal divergence from reference answers, as long as semantic proximity is maintained. This approach aligns well with the principles of natural language processing, which emphasize semantic understanding.

### *4.8. Examples of computational problems*

In our evaluation, as illustrated in Table 10, we present four distinct examples of computational problem cases, each designed to test the mathematical problem-solving abilities of the models. These problems vary in complexity and type, offering a comprehensive assessment of the models' computational prowess. For each problem, we set specific prompts and format output requirements to standardize the testing procedure and ensure comparability across different models.

The evaluation process for each computational problem is conducted in several systematic steps:

**Extraction of Final Answer:** Initially, we extract the final answer from the model's response. This step is crucial as it focuses on isolating the core numerical or symbolic output that the model has generated in response to the computational problem.

**Standardization of the Answer:** Once the final answer is extracted, we undertake a standardization process. This involves removing any spaces, superfluous symbols, or extraneous characters that do not contribute to the mathematical validity of the answer. The purpose of this step is to ensure that the answers can be uniformly evaluated, irrespective of minor variations in formatting or presentation that might be present in the models' responses.

**Comparison with Reference Answer:** The standardized final answer is then compared with the reference final answer. The reference answers are predetermined solutions that are known to be correct. This comparison is critical to determine the accuracy of the model's response.

**Gscore Calculation for Correct Answers:** If the model's standardized final answer completely matches the reference final answer, it indicates that the problem has been correctly solved. In such cases, the model is awarded a Gscore of 1, denoting full marks for accuracy and correctness in problem-solving.

**Gscore Calculation for Incorrect Answers:** In instances where the model's answer does not match the reference answer, we proceed to evaluate the problem-solving process. We calculate the Chrf score, which is a character-level evaluation metric, of the model's entire problem-solving process against the reference process. The Chrf score thus reflects how closely the model's approach and intermediate steps align with the standard method. This score is then multiplied by a factor of 0.3 to obtain the final Gscore for that problem. This reduced weighting underscores the importance of not just the process but also the accuracy of the final answer.

Through this meticulous and structured evaluation process, we aim to provide a robust and fair assessment of each model's ability to tackle and accurately solve mathematical problems. This methodology not only tests the models' computational accuracy but also their adherence to logical problem-solving methods, offering a comprehensive insight into their capabilities in mathematical reasoning and computation.

### *4.9. Correlation between the results of human evaluation and Gscore*

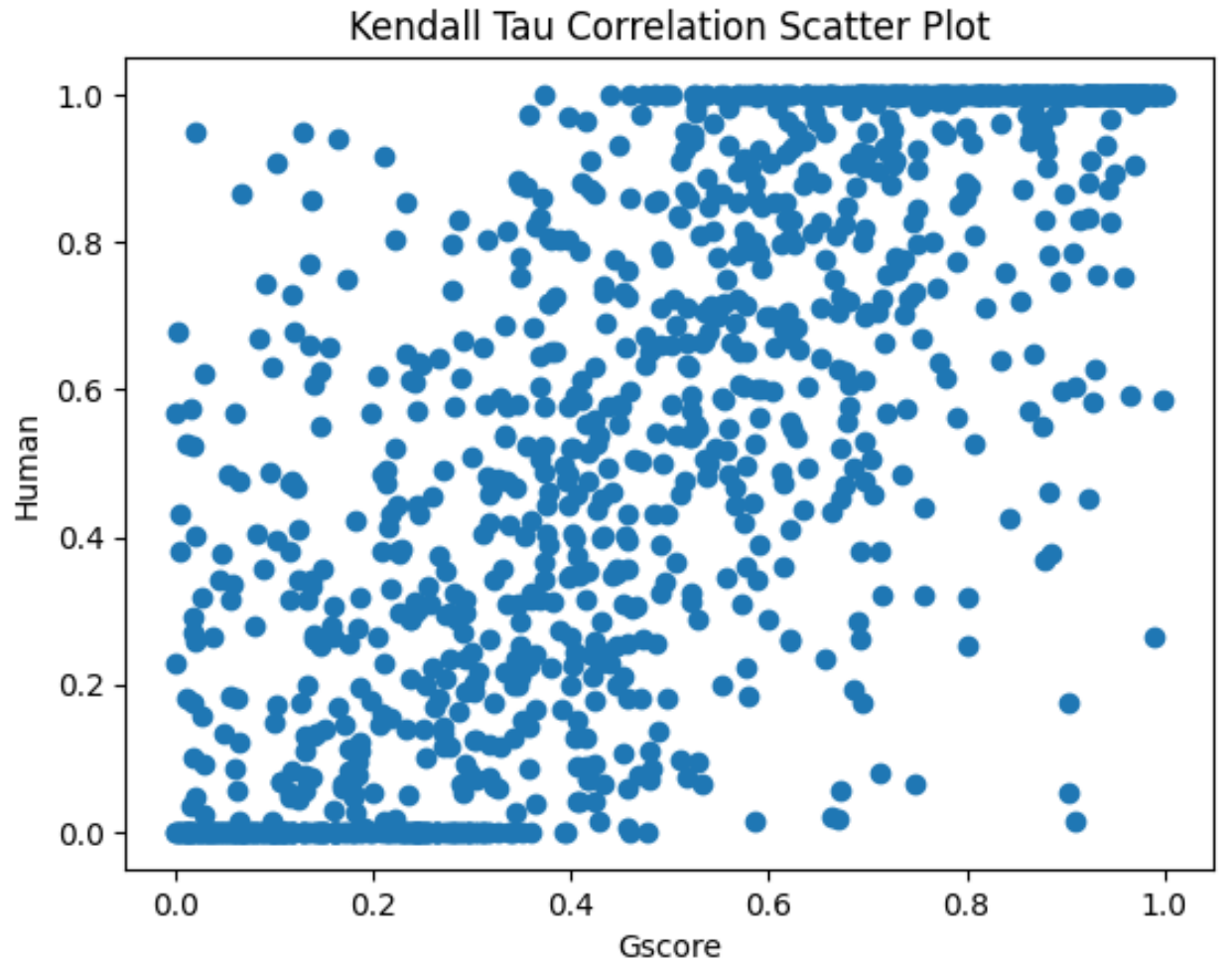

**Fig. 2.** Kendall Tau Correlation.

The correlation results are presented as follows, with the scatter plot illustrated in Figure 2.

Kendall Tau correlation: 0.6137523339513649

P-value: 1.5991267354009107e-215

To evaluate the correlation between Gscore and human assessment, we conducted a study using 11,000 questions to test 10 large language models, resulting in corresponding Gscores. To assess the correlation, we extracted a subset of 1,100 question-and-answer pairs from the total set. These pairs were derived from 55 distinct subjects and responses generated by the 10 different models. Subsequently, we invited human experts to score these 1,100 pairs. The final step involved calculating the correlation between the Gscores and the expert human evaluations.

This methodology ensures a comprehensive and representative evaluation of the models' performance across a diverse range of subjects. The involvement of human experts in scoring provides a robust benchmark for assessing the validity and reliability of the Gscores.

The outline of the sampling scheme is as follows:

**Stratified Random Sampling**: Given the diversity of subjects, with 55 different ones in total, stratification by subject should be first conducted to ensure representativeness within each stratum. Within each subject, random sampling will be employed to select questions, ensuring variety and representativeness in the sample.

**Table 10**
Examples of computational problems.

<table>
<tr><td rowspan="9">Example 1</td><td>Question</td><td colspan="2">以下是小学数学科目的问题，请以“解：”开始给出解题过程，并在解题过程的最后换行，在最后一行以“最终答案：”开头，按顺序给出数值及其单位，采用英文逗号分割，例如“最终答案：1元,1次,1公顷,1人”。<br>有甲乙丙三袋化肥，甲乙两袋共重32千克，乙丙两袋共重30千克，甲丙两袋共重22千克，求三袋化肥各重多少千克。</td></tr>
<tr><td>Model output</td><td colspan="2">解：设甲乙丙三袋化肥的重量分别为x千克，y千克，z千克。<br>根据题目，我们可以得到以下三个方程：<br>x + y = 32<br>y + z = 30<br>x + z = 22<br>我们可以先将第一个方程和第三个方程相加，得到：2x + y + z = 54<br>然后将这个方程和第二个方程相减，得到：x = 24<br>将x = 24代入第一个方程，得到：y = 8<br>将y = 8代入第二个方程，得到：z = 22<br>最终答案：24千克,8千克,22千克。</td></tr>
<tr><td>Extracted final answer</td><td colspan="2">24千克,8千克,22千克。</td></tr>
<tr><td>Standardized final answer</td><td colspan="2">24千克,8千克,22千克</td></tr>
<tr><td>Reference</td><td colspan="2">解：<br>甲乙两袋、乙丙两袋都含有乙，从中可以看出甲比丙多（32－30）＝2千克，且甲是大数，丙是小数。由此可知<br>甲袋化肥重量＝（22＋2）÷2＝12（千克）<br>丙袋化肥重量＝（22－2）÷2＝10（千克）<br>乙袋化肥重量＝32－12＝20（千克）<br>答：甲袋化肥重12千克，乙袋化肥重20千克，丙袋化肥重10千克。</td></tr>
<tr><td>Reference final answer</td><td colspan="2">24千克,8千克,22千克</td></tr>
<tr><td colspan="2">Gscore</td><td>Human</td></tr>
<tr><td colspan="2">1</td><td>1</td></tr>
</table>

<table>
<tr><td rowspan="9">Example 2</td><td>Question</td><td colspan="2">以下是小学数学科目的问题，请以“解：”开始给出解题过程，并在解题过程的最后换行，在最后一行以“最终答案：”开头，按顺序给出数值及其单位，采用英文逗号分割，例如“最终答案：1元,1次,1公顷,1人”。<br>学校把植树560棵的任务按人数分配给五年级三个班，已知一班有47人，二班有48人，三班有45人，三个班各植树多少棵？</td></tr>
<tr><td>Model output</td><td colspan="2">解：首先，我们需要知道五年级总共有多少人。这可以通过将三个班的人数相加得到，即47人+48人+45人=140人。<br>然后，我们可以通过将总的植树任务数除以总人数，得到每个人需要植树的数量。即560棵/140人=4棵/人。<br>最后，我们可以通过将每个班的人数乘以每个人需要植树的数量，得到每个班需要植树的数量。即一班需要植树47人*4棵/人=188棵，二班需要植树48人*4棵/人=192棵，三班需要植树45人*4棵/人=180棵。<br>最终答案：188棵,192棵,180棵。</td></tr>
<tr><td>Extracted final answer</td><td colspan="2">188棵,192棵,180棵。</td></tr>
<tr><td>Standardized final answer</td><td colspan="2">188棵,192棵,180棵</td></tr>
<tr><td>Reference</td><td colspan="2">解：<br>总份数为47＋48＋45＝140<br>一班植树560×47/140＝188（棵）<br>二班植树560×48/140＝192（棵）<br>三班植树560×45/140＝180（棵）<br>答：一、二、三班分别植树188棵、192棵、180棵。</td></tr>
<tr><td>Reference final answer</td><td colspan="2">188棵,192棵,180棵</td></tr>
<tr><td colspan="2">Gscore</td><td>Human</td></tr>
<tr><td colspan="2">1</td><td>1</td></tr>
</table>

<table>
<tr><td rowspan="9">Example 3</td><td>Question</td><td colspan="2">以下是高中数学科目的问题，请使用latex语法给出解题过程，并在解题过程的最后换行，在最后一行以“最终答案：”开头，根据不同的题目类型按照latex语法给出数值、表达式、导数、积分、方程的根。导数根据题目表述采用latex语法按照y'或者f'(x)表示。如果方程的一个未知数有多个解，答案采用形如“x=1或x=-3”的方式表示。如果方程有多个未知数，答案采用形如“x=1,y=-3,z=5”的方式表示，用英文逗号分隔。以下为需要解答的题目：<br>计算 \left(\frac{1}{3}\right)^{-1}-\log _{2} 8+\left(0.5^{-2}-2\right) \times\left(\frac{27}{8}\right)^{\frac{2}{3}} 的值</td></tr>
<tr><td>Model output</td><td colspan="2">首先，我们可以将每个部分分开计算。<br>对于 \left(\frac{1}{3}\right)^{-1}，我们可以得到：<br>\left(\frac{1}{3}\right)^{-1} = 3<br>对于 \log _{2} 8，我们可以得到：\log _{2} 8 = 3<br>对于 0.5^{-2}-2，我们可以得到：0.5^{-2}-2 = 4-2 = 2<br>对于 \left(\frac{27}{8}\right)^{\frac{2}{3}}，我们可以得到：\left(\frac{27}{8}\right)^{\frac{2}{3}} = \left(\frac{3}{2}\right)^2 = \frac{9}{4}<br>然后，我们可以将这些结果代入原式，得到：3 - 3 + 2 \times \frac{9}{4} = 3 - 3 + \frac{9}{2} = 0 + \frac{9}{2} = \frac{9}{2}<br>所以，原式的值为 \frac{9}{2}。<br>最终答案：\frac{9}{2}</td></tr>
<tr><td>Extracted final answer</td><td colspan="2">\frac{9}{2}</td></tr>
<tr><td>Standardized final answer</td><td colspan="2">\frac{9}{2}</td></tr>
<tr><td>Reference</td><td colspan="2">\left(\frac{1}{3}\right)^{-1}-\log _{2} 8+\left(0.5^{-2}-2\right) \times\left(\frac{27}{8}\right)^{\frac{2}{3}}\begin{array}{l}=3-3 \log _{2} 2+(4-2) \times \frac{9}{4} \\ =\frac{9}{2} .\end{array}</td></tr>
<tr><td>Reference final answer</td><td colspan="2">\frac{9}{2}</td></tr>
<tr><td colspan="2">Gscore</td><td>Human</td></tr>
<tr><td colspan="2">1</td><td>1</td></tr>
</table>

<table>
<tr><td rowspan="9">Example 4</td><td>Question</td><td colspan="2">以下是大学数学科目的问题，请使用latex语法给出解题过程，并在解题过程的最后换行，在最后一行以“最终答案：”开头，根据不同的题目类型按照latex语法给出数值、表达式、导数、积分、方程的根。导数根据题目表述采用latex语法按照y'或者f'(x)表示。如果方程的一个未知数有多个解，答案采用形如“x=1或x=-3”的方式表示。如果方程有多个未知数，答案采用形如“x=1,y=-3,z=5”的方式表示，用英文逗号分隔。以下为需要解答的题目：<br>求下列不定积分：\int \frac{d x}{x^{2}}</td></tr>
<tr><td>Model output</td><td colspan="2">对于这个不定积分，我们可以直接使用基本积分公式进行求解。<br>基本积分公式中，$\int \frac{1}{x} dx = \ln |x| + C$，所以$\int \frac{1}{x^2} dx = \int x^{-2} dx$。<br>根据幂函数的积分公式，$\int x^n dx = \frac{1}{n+1}x^{n+1} + C$，当$n \neq -1$时。<br>所以，我们可以将$x^{-2}$看作$x^{n}$，其中$n=-2$，代入公式求解。<br>\[<br>\int x^{-2} dx = \frac{1}{-2+1}x^{-2+1} + C = -x^{-1} + C = -\frac{1}{x} + C<br>\]<br>最终答案：$-\frac{1}{x} + C$</td></tr>
<tr><td>Extracted final answer</td><td colspan="2">$-\frac{1}{x} + C$</td></tr>
<tr><td>Standardized final answer</td><td colspan="2">-\frac{1}{x}+C</td></tr>
<tr><td>Reference</td><td colspan="2">\int \frac{\mathrm{d} x}{x^{2}}=\int x^{-2} \mathrm{~d} x=\frac{1}{-2+1} x^{-2+1}+C=-\frac{1}{x}+C</td></tr>
<tr><td>Reference final answer</td><td colspan="2">-\frac{1}{x}+C</td></tr>
<tr><td colspan="2">Gscore</td><td>Human</td></tr>
<tr><td colspan="2">1</td><td>1</td></tr>
</table>

**Model Coverage Assurance**: From the responses provided by the 10 large models, for each selected question, a response from a randomly chosen model will be included. This approach ensures that the analysis covers responses from all participating models, enhancing the comprehensiveness of the evaluation.

**Balanced Subject Distribution:** Ensure that the proportion of each subject in the 1100 sampled questions is roughly equivalent to its proportion in the total pool of 11000 questions. This balance maintains the representativeness of the sample, avoiding overrepresentation or neglect of certain subjects.

### *4.10. Why adopt fixed prompts?*

Consistency and Comparability: Fixed prompts ensure the consistency of the evaluation process, allowing for direct comparison of results between different models. This standardization is crucial for fairly assessing the performance of various models.

Control of Variables: In scientific research, controlling variables is essential. By using fixed prompts, researchers can eliminate performance variations caused by differing prompts, thereby more accurately assessing the model's inherent capabilities.

Reproducibility: Fixed prompts enhance the reproducibility of experiments. Other researchers can replicate the experiments using the same prompt words and validate or compare their results.

Simplification of the Evaluation Process: The use of fixed prompts simplifies the evaluation process, making model assessment more accessible and comprehensible, especially for non-expert users.

## 5. Discussion

### *5.1. Human-level test benchmarks*

The survey "A Survey of Large Language Models" (Zhao et al., 2023) identifies three primary methods for evaluating Large Language Models (LLMs): the benchmark-based approach, the human-based approach, and the model-based approach. The CG-Eval falls under the category of a benchmark-based approach. This method utilizes human-level test benchmarks designed to assess the overall capabilities of LLMs using questions typically intended for human examination. Examples of such benchmarks include MMCU (Zeng, 2023), C-Eval (Huang et al., 2023), M3KE (Liu et al., 2023) , GAOKAO-Bench (Zhang et al., 2023), Xiezhi (獬豸) (Gu et al., 2023) and CMMLU (Li et al., 2023). These benchmarks cover a broad spectrum of topics, varying levels of difficulty, and multiple languages, ensuring a thorough evaluation of the LLMs' general performance.

In conducting benchmark evaluations, each question is first transformed into a prompt that the LLMs respond to, generating textual output. This output is then analyzed using predefined human-crafted rules to extract the LLMs' predicted answers. The effectiveness and accuracy of the LLMs are quantitatively measured by comparing these predicted answers against the actual, correct answers. This comparison employs standard metrics such as accuracy. The evaluation process itself can be executed in different scenarios, including few-shot and zero-shot settings. These varying settings can influence the outcomes of the evaluation, potentially affecting the performance results and rankings of the LLMs.

### *5.2. Pros and Cons of Benchmark-based approach*

The benchmarks used to evaluate large language models (LLMs) typically include a robust number of test samples that effectively gauge their fundamental capabilities. This evaluation process is largely automated, making it a convenient method for conducting experiments across various LLMs. This is particularly beneficial for tracking the performance of models at different checkpoints during the pre-training phase. Nonetheless, it's important to recognize that LLMs' performance can be significantly influenced by several factors in the evaluation setup. These include the style of the question prompts, whether the tests are conducted in a zero-shot or few-shot manner, and the methods used for parsing answers. Therefore, when executing these evaluations, it's crucial to consider these potential variables. Additionally, the specific settings used in the evaluation must be clearly noted alongside the results.

Another critical issue to consider is data contamination. This occurs when the test data, or material closely related to it, has already been included in the pre-training datasets of the LLMs. With the increasing trend of incorporating vast amounts of open data into the development of LLMs, data contamination has become a more prominent concern that needs to be addressed in the evaluation process.

### *5.3. Possible Future Improvements*

Considering the advancements in the field, we are contemplating the adoption of Para-Ref (Tang et al., 2023), an innovative method, for future enhancements of our Gscore metric. The Para-Ref approach, which focuses on expanding the range and depth of evaluation benchmarks, achieves this by increasing the quantity of reference texts. It utilizes large language models (LLMs) to paraphrase existing reference materials into various high-quality expressions.

The potential integration of Para-Ref (Tang et al., 2023) into the Gscore evaluation framework is envisioned to tackle a critical issue in generative model assessment: the limitations arising from reliance on a single or limited set of reference texts. By augmenting the pool of reference texts through paraphrasing, we aim to enable Gscore to encompass a more extensive array of linguistic expressions and subtleties. This approach would facilitate a more comprehensive and nuanced evaluation of models' generative capabilities.

Drawing on empirical evidence from fundamental Natural Language Generation (NLG) tasks, including machine translation, text summarization, and image captioning, the effectiveness of such an approach is evident. The application of Para-Ref (Tang et al., 2023) has shown considerable promise in enhancing the alignment of automatic evaluation metrics with human judgment, evidenced by significant improvements in correlation ratios. This potential enhancement to Gscore is particularly crucial, as it would heighten the metric's sensitivity and alignment with human evaluative standards, thereby offering a more accurate and reliable assessment of generative models. We are optimistic that the future incorporation of Para-Ref into our evaluation process will significantly advance the precision and reliability of the Gscore metric.

### *5.4. Conclusion*

The landscape of large-scale language models is undergoing rapid and dynamic advancements. Despite the recent influx of new evaluation benchmarks for these models, a notable gap persists in both English and Chinese sectors, particularly in metrics tailored for assessing text generation capabilities. Addressing this gap, our work introduces CG-Eval, a pioneering benchmark specifically designed to evaluate Chinese large-scale models. CG-Eval stands out in its comprehensive ability to assess not just the accuracy and relevance of responses across diverse domains but also the models' proficiency in complex mathematical computations. This makes it an all-encompassing benchmark, pivotal for advancing the field of Chinese text generation.

A significant achievement of CG-Eval is the full automation of the evaluation process, ensuring efficiency and objectivity. This automation enables rapid acquisition and analysis of results, eliminating the need for manual intervention and thereby enhancing the scalability and reliability of the evaluation.

However, we recognize that CG-Eval is not without its limitations. Currently, it does not extend to fully automating the evaluation of aspects such as role-playing, multi-turn dialogues, creative generation, casual conversation, and logical reasoning in Chinese large-scale models. These areas represent the next frontier for our research. We are optimistic that with continued innovation and development, these challenges will soon be overcome. Our commitment to evolving CG-Eval aligns with the broader objective of advancing the field of AI and language modeling. We anticipate that our contributions will not only bolster current model evaluations but also inspire future research, paving the way for more sophisticated and nuanced language models.

In conclusion, our work with CG-Eval marks a significant step forward in the realm of language model evaluation. It lays a solid foundation for future advancements, promising to bridge existing gaps and expand the horizons of what these remarkable models can achieve.

## 6. Author Contributions

Evaluation System Design and Validation: Hui Zeng.
Data Collection and Annotation: Chen Sun.

Data Processing: Hui Zeng.

Evaluation: Meng Hao was responsible for testing models accessible through web interfaces and APIs. Jingyuan Xue focused on testing models accessible through their weights.

Website and Submission System: Bin Ning and Hui Zeng built the website and the online submission system.

Paper Writing: Hui Zeng wrote the main content of this paper.

Illustration Creation and Website Logo Design: Meng Hao, Na Zhang.